\documentclass[runningheads]{llncs}
\usepackage[T1]{fontenc}
\usepackage{graphicx}
\usepackage{booktabs}
\usepackage{pifont}
\usepackage[accsupp]{axessibility}  

\usepackage[utf8]{inputenc}
\usepackage{soul}
\usepackage{pifont}
\usepackage{amssymb}
\usepackage{bm}
\usepackage{color}
\usepackage[dvipsnames]{xcolor}
\usepackage{gensymb}
\usepackage{orcidlink}
\usepackage{amsmath}

\begin{document}

\title{SalFoM: Dynamic Saliency Prediction with Video Foundation Models} 

\titlerunning{Dynamic Saliency Prediction with VFMs}

\author{
  Morteza Moradi\inst{1} \and
  Mohammad Moradi\inst{1} \and
  Francesco Rundo\inst{2}  \and 
  Concetto Spampinato\inst{1} \and
  Ali Borji\inst{3}\thanks{ Equal supervision}    \and
  Simone Palazzo\inst{1}\inst{\star}
}
\authorrunning{Moradi et al.}

\institute{University of Catania, Catania, Italy \\
\email{\{morteza.moradi, mohammad.moradi\}@phd.unict.it,\\\{concetto.spampinato, simone.palazzo\}@unict.it} \and
STMicrolectronics, ADG Central R \& D, Catania, Italy\\
\email{francesco.rundo@st.com}\and
Quintic AI, San Francisco, CA, USA\\
\email{aliborji@gmail.com}}

\maketitle              

\begin{abstract}
Recent advancements in video saliency prediction (VSP) have shown promising performance compared to the human visual system, whose emulation is the primary goal of VSP. However, current state-of-the-art models employ spatio-temporal transformers trained on limited amounts of data, hindering generalizability adaptation to downstream tasks. The benefits of vision foundation models present a potential solution to improve the VSP process. However, adapting image foundation models to the video domain presents significant challenges in modeling scene dynamics and capturing temporal information. To address these challenges, and as the first initiative to design a VSP model based on video foundation models, we introduce SalFoM, a novel encoder-decoder video transformer architecture. Our model employs UnMasked Teacher (UMT) as feature extractor and presents a heterogeneous decoder which features a locality-aware spatio-temporal transformer and integrates local and global spatio-temporal information from various perspectives to produce the final saliency map. Our qualitative and quantitative experiments on the challenging VSP benchmark datasets of DHF1K, Hollywood-2 and UCF-Sports demonstrate the superiority of our proposed model in comparison with the state-of-the-art methods.
\end{abstract}

\section{Introduction}
\label{sec:intro}

Video saliency prediction, i.e., modeling the focus of attention of the human visual system when observing a dynamic scene, has increasingly gained attention in recent years~\cite{TMFI}, also due to the growing demand for video content understanding and analysis across various application domains.  
Numerous deep learning-based strategies have been explored to enhance the accuracy and performance of these methods, aiming to reach human-level scene recognition. Among the different approaches, state-of-the-art methods that have shown the best results to date utilize spatio-temporal transformers as encoder parts. However, since these networks are often not trained on massive datasets, their generalizability and adaptability for downstream tasks are limited.

The emergence of foundation models \cite{bommasani2021opportunities} offers a solution to this fundamental challenge, as these models are trained on vast and diverse datasets, encompassing a large variability and gaining high generalizability without the need for re-training.
In line with the core goal of foundation models to achieve human-like intelligence and understanding, viable solutions for video saliency prediction can leverage the capabilities of video foundation models (VFMs).

Most of current designs of VFMs are built upon robust image foundation models (IFMs), such as CLIP-ViP~\cite{xue2022clip}, based on CLIP~\cite{clip}. Although this approach is cost-effective, as it builds on pre-trained static features, it also presents significant challenges due to the nature of IFMs, that overlook temporal and motion-related features. Therefore, such models may not be fully suitable for adaptation in various video understanding tasks, including video saliency prediction. To fill this gap, in this work we employ UnMasked Teacher (UMT)~\cite{UMT}, a pure video foundation model that retains spatio-temporal features of video content, aiming to handle various video-centric tasks such as video-text retrieval and action recognition.

In this work, we design a novel dynamic saliency prediction model that is empowered by a video foundation model based encoder. To fully exploit the expressive power of spatio-temporal representations extracted by the encoder, we introduce a decoder architecture that is composed of three different intermediate branches, each of them reconstructing features from different perspectives.
More specifically, one of the branches employs spatio-temporal transformers~\cite{liu2022video} to extract long-range spatio-temporal relationships, operating at the same resolution of the encoded features; the second branch extracts local spatio-temporal representations, gradually reducing the temporal resolution and compensating for the lack of global information by a feature fusion mechanism with the first branch; the final branch focuses instead on the spatial relations between scene elements, collapsing the temporal dimension and producing high-resolution features to guide the synthesis of the output saliency map, while at the same time incorporating information from the previous two branches. We conduct extensive quantitative and qualitative experiments on the standard VSP dataset, namely DHF1K, Hollywood-2 and UCF-Sports; our findings uncover the superiority of our model over the state-of-the-art models.

To summarize our contributions:
\begin{itemize}
\item We propose the first video saliency prediction model based on a pure video foundation model to capture spatio-temporal features of video content.
\item We introduce a novel heterogeneous decoder network that employs locality-aware
spatio-temporal attention to better process encoder features.
\item We show the superiority of the performance of our model on the most challenging video saliency dataset, DHF1K, compared to the state-of-the-art VSP models.
\end{itemize}

\section{Related Work}

\subsection{Video Saliency Prediction}

Deep learning-based video saliency prediction, as explored in~\cite{wang2019revisiting}, has recently become a prominent method for modeling human gaze in dynamic scenes. The primary goal of video saliency prediction (VSP) is to mimic the human visual system and create patterns of attention allocation for video frames. One practical application of these models is predicting a driver's focus of attention in traffic scenarios~\cite{xia2019predicting}, which is crucial for decision-making processes.

One of the most effective VSP models in the pre-transformer era is STSANet \cite{wang2021spatio}, which addresses the challenge of understanding long-range temporal relationships in videos. It features a 3D fully convolutional network as its backbone and is structured as a four-branch network. The network utilizes spatio-temporal self-attention (STSA) modules to capture spatio-temporal dependencies and employs attentional multi-scale fusion modules to integrate the extracted features. In the audio-visual VSP domain, TSPF-Net~\cite{chang2021temporal} addresses saliency modeling by designing a feature pyramid network that integrates scale, space and time. The network hierarchically decoded features at various levels of the pyramid, considering the impact of spatial and temporal features at different scales. Following a different approach, HD2S~\cite{HD2S} constructs multiple intermediate saliency maps at various levels of abstraction, which are then integrated to produce the final saliency map. This design aims to incorporate both general and data-specific features into the saliency prediction process.

VSFT~\cite{9770033} is the first to employ transformer architectures in VSP, focusing on forecasting saliency for unseen future frames. Unlike the CNN-based models mentioned before, VSFT uses a self-attention mechanism to capture both short- and long-range relationships between video frames. Its decoder combines the encoded features using proposed cross-attention guidance blocks (CAGB) to capture spatio-temporal correlations. Another transformer-based model, THTD-Net~\cite{THTD}, stands out as a lightweight solution for VSP, where most of the temporal information is processed in the decoding stage. Opting for a single-branch decoder without reducing the encoded features before decoding, the model manages to use fewer parameters, compared to attention-based or multi-branch approaches. Despite its relative simplicity, THTD-Net demonstrated performance on par with state-of-the-art solutions. Another transformer-based model, TMFI-Net~\cite{TMFI}, consists of a semantic-guided encoder and a hierarchical decoder. The encoder captures spatio-temporal features and provides semantic contextual information, integrating this information in a top-down manner using a feature pyramid structure. Before decoding, a multi-dimensional attention (MA) module is used to enhance the spatio-temporal features. 

In this work, we present the first dynamic saliency prediction model that is empowered by a video foundation model for feature encoding. Moreover, unlike other approaches, where different decoding branches generally focus on varying spatial scales with a late feature fusion mechanism (e.g.,~\cite{HD2S}), our model processes and combines encoded features from different temporal viewpoints to gradually reconstruct crucial features for the ultimate saliency map prediction. 

\subsection{Video Foundation Models}

Although foundation models for vision~\cite{awais2023foundational} have become increasingly prominent in recent years, the majority of published works and practical efforts have focused on image foundation models, as seen in~\cite{yuan2021florence,yu2022coca}. The growing need to understand and analyze the pervasive and continuously-generated video content, ranging from social media and sports videos to traffic and surveillance footage, has spurred the research community to harness the power of foundation models for video-based tasks. However, the development of such models faces two significant challenges: the scarcity of a large and diverse video dataset and the substantial computational costs involved to train such models.

To overcome these obstacles, research has shifted towards the creation of video foundation models (VFMs) that build upon image foundation models (IFMs). Leveraging the strength and versatility of established large image models, such as CLIP~\cite{clip}, several vision models have been modified to use IFMs for addressing downstream video-related tasks. InternVideo~\cite{wang2022internvideo} proposes a general video foundation model by introducing the concept of a unified video representation. This model utilizes UniformerV2~\cite{li2023uniformerv2} as its encoder and is built upon CLIP. In an effort to devise an efficient approach for translating masked modeling to videos, VideoMAE~\cite{tong2022videomae} presents a straightforward design with reduced computational costs by employing an asymmetric encoder-decoder structure. Building upon this, VideoMAEv2~\cite{wang2023videomae} adopts a dual masking strategy to enhance the original model, making it more efficient for pre-training.

Although IFMs have facilitated the development of large video models in certain respects, such models still struggle with handling the temporal dynamics that are inherent to video data. To deal with such issues, UMT, as the first attempt to design a native large video model, opted for an innovative strategy to both make the training process efficient and preserve temporal information. It utilizes CLIP-ViT as an Unmasked Teacher to train a vanilla spatio-temporal ViT from scratch for masked video modeling. UMT preserves the spatial architecture of the teacher model for processing each frame individually. Furthermore, it leverages spatio-temporal attention to facilitate interaction among the unmasked tokens. This strategy not only enables handling limited data scale for video understanding tasks but also accelerates convergence and significantly enhances the model's capability to capture temporal information across frames.

\section{Methodology}

\begin{figure*}[t]
    \centering
    \includegraphics[width=1\textwidth]{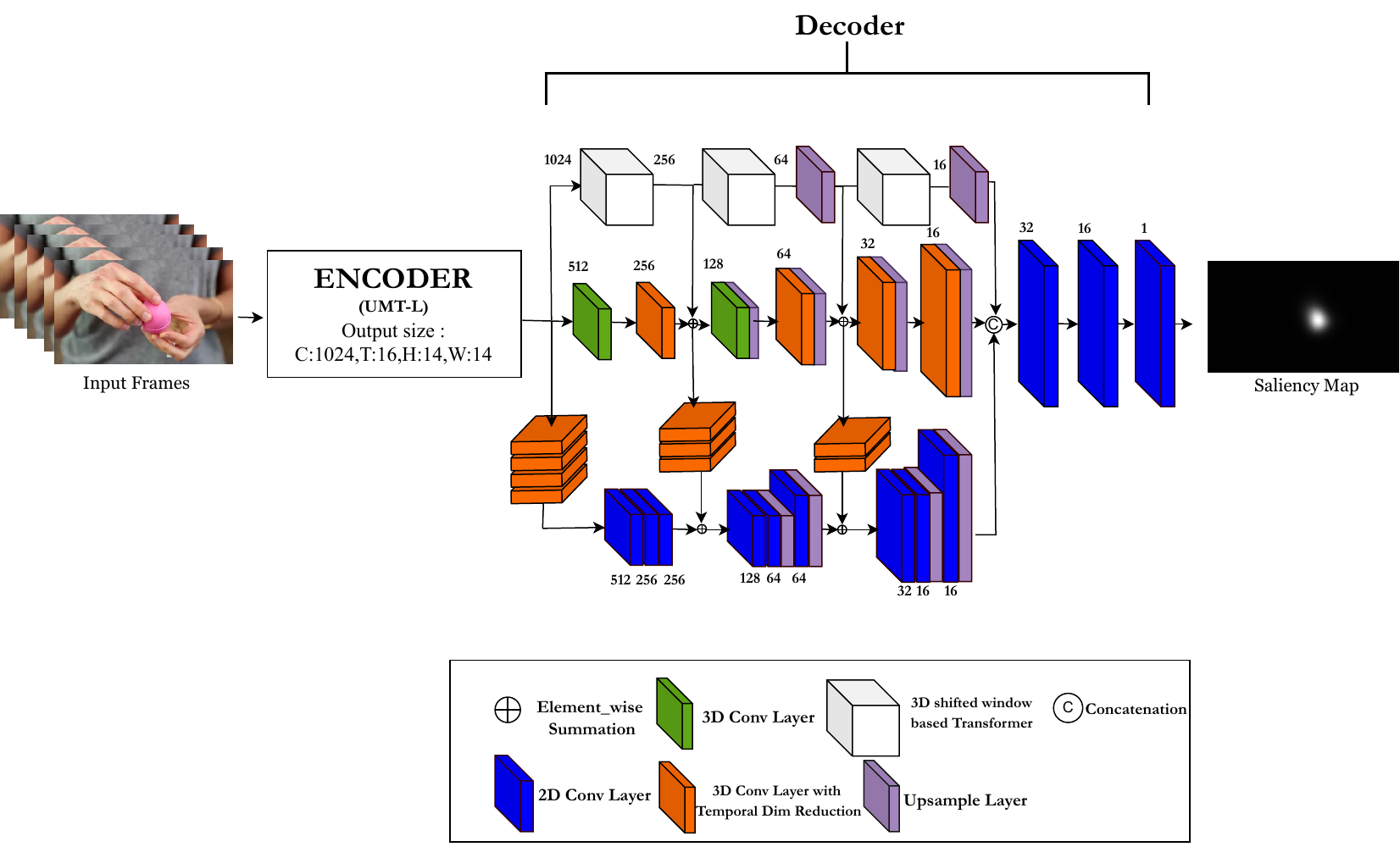}
    \caption{Summary of the proposed dynamic saliency prediction framework. The model utilizes a video foundation model called UMT-L as its encoder, while the decoding phase comprises three different intermediate branches aimed at progressively refining and reconstructing essential features for producing the ultimate saliency map. In the end, all intermediate features are combined to generate the final saliency map.}
    \label{network}
\end{figure*}

The progressive development of vision foundation models, that aims at building models with high generalizability for being adapted to various downstream tasks, motivated us to design a dynamic saliency prediction model based on a video foundation model (Figure~\ref{network}), namely Unmasked Teacher (UMT), as the encoder part of our network. To the best of our knowledge, this is the earliest work that incorporates a purely spatio-temporal foundation model into the video saliency prediction's workflow. Moreover, we designed a heterogeneous multi-branch decoder, consisting of both dynamic and static branches, that is intended to include spatial and temporal information in the process of generating saliency map for an input video frame from different perspectives.

\subsection{Video Foundation Model--based Feature Encoder}

In our architecture, we employ UMT~\cite{UMT} as our network’s encoder. In contrast to other video foundation models, that are directly adapting image foundation models, UMT is trained by using CLIP-ViT \cite{clip} as Unmasked Teacher to train a vanilla spatio-temporal ViT from scratch. Training is carried out using a self-supervised teacher-student knowledge distillation approach, by masking out and predicting most of the video tokens with low semantics and aligning the unmasked tokens with a linear projection to the corresponding ones from the teacher, in order to handle the limited data scale and effectively utilize the video data. Spatio-temporal attention~\cite{DBLP:conf/icml/BertasiusWT21} is exploited to facilitate the interaction between all the unmasked tokens with each other. The architecture does not use temporal downsampling, which ensures that tokens can be aligned frame by frame.

In our framework, we employ the pre-trained large version of the UMT, referred to as UMT-L, in the feature encoding stage. After removing the classification head, the model can be treated as a feature extractor, which receives an input video \(\mathbf{V} \in \mathbb{R}^{T \times H \times W \times 3}\) and provides features \( \mathbf{F} \in \mathbb{R}^{T \times h \times w \times f} \), with \(h = \frac{H}{16}\), \(w = \frac{W}{16}\) and \(f = 1024\). More details about model design are provided in the supplementary materials.

\subsection{Multiperspective Heterogeneous Decoder}

The design of the proposed multiperspective heterogeneous decoder is based on the principle of capturing and integrating diverse aspects of spatio-temporal information encoded by the VFM feature encoder. The rationale behind this approach is to ensure that the final saliency map is a comprehensive representation of the most salient features in both space and time, which is crucial for attention modeling in videos. To this aim, the proposed decoder is designed according to the following insights and principles:
\begin{itemize}
\item \emph{Gradual temporal dimension reduction}: Inspired by existing strategies, that shows the benefits of keeping the temporal resolution as close as possible to the original input~\cite{THTD}, our approach avoids abrupt loss of temporal information by gradually reducing the temporal dimension.
\item \emph{Channel dimension reduction}: We exploit the strong expressive power of features provided by the VFM encoder, and hypothesize that the objective of the decoder is to find a suitable feature analysis and interaction modality, rather than \emph{extracting} more complex features. Hence, we encourage our model to distill the essential features by reducing the channel dimension from a high-dimensional space to a more compact representation, facilitating efficient computation and potentially improving generalization.
\item \emph{Heterogeneous spatio-temporal feature decoding}: The first intermediate branch of our decoder network (Transformer-based Complementary Feature Extraction, TCFE) is designed to capture spatio-temporal relationships and encode them into feature maps. The second branch (Dynamic Feature Decoding, DFD) focuses, instead, on maintaining temporally-rich information and extracting detailed local features. While the locality of this operation limits the discovery of useful long-range spatio-temporal patterns, it allows to gradually increase and recover the original input resolution. The third branch (Static Feature Decoding, SFD), finally, abstracts the temporal effects to focus on spatial information, recognizing that not all temporal information is equally relevant for saliency prediction.
\item \emph{Feature Fusion}: The final feature fusion stage integrates features from all branches, allowing the network to leverage the diverse perspectives captured by each branch. This integrated representation is then processed through 2D convolutional layers to produce the final saliency map.
\end{itemize}

Formally, Let \(\mathbf{F} \in \mathbb{R}^{t \times h\times w \times f}\) be the set of spatio-temporal features obtained from the transformer feature extractor, with \(t\), \(h\), \(w\), and \(f\) representing the temporal, height, width, and feature dimensions, respectively. The objective is to map \(\mathbf{F}\) to a saliency map \(\mathbf{S} \in \mathbb{R}^{H \times W}\), where \(H\) and \(W\) are the dimensions of the original frame.

Let us model the first branch of the decoder, i.e. the TCFE subnetwork, as a sequence of $N$ layers, producing features \( \bm{\Theta} = \left\{ \bm{\theta_1}, \dots, \bm{\theta_N} \right\} \). At this perspective of the mode, features \(\bm{\Theta}\) are intended to extract long-range spatio-temporal relationships, operating at the same resolution as input features \(\mathbf{F}\), acting only on the channel dimension. For this reason, we do not let \(\bm{\Theta}\) be affected by higher-resolution features extracted by other branches, as the advantage of the latter, i.e., the higher detail, would be inevitably lost due to downsampling to the \(t \times h \times w\) resolution. 

The second branch of the decoder, DFD, is made up of the same number of layers as the first, and extracts features \( \bm{\Phi} = \left\{ \bm{\phi}_1, \dots, \bm{\phi}_N \right\} \). This branch is dedicated to extracting local spatio-temporal features information, increasing the spatial resolution to recover details while gradually reducing the temporal dimension to retain as much of the video dynamics as possible. In order to compensate for the lack of global analysis in this branch, we integrated features from the first branch at corresponding position of the layer cascade, through proper upsampling. In detail, we compute feature \( \bm{\phi}_i \), with \( i > 1\), as follows:
\begin{equation}
\bm{\phi}_i = f_i \left( \bm{\phi}_{i-1} \right) \oplus \sigma_i \left( \bm{\theta}_i \right) ,
\end{equation}
where \( f_i \) is the transformation applied at the $i$-th layer of the branch, and \( \sigma_i \) is the appropriate down-sampling function.

The third branch of the decoder, SFD, extracts features \( \bm{\Gamma} = \left\{ \bm{\gamma}_1, \dots, \bm{\gamma}_N \right\} \); the main property of this branch is that it collapses the temporal dimension of its input features, summarizing them into a single channel and focusing instead on spatial relations between scene elements. To this aim, each layer \(g_i\) receives a temporally-collapsed version of the input frames, using a learned transformation \(\tau_i\), and produces upscaled features, with the same resolution as the corresponding ones from \(\bm{\Phi}\). In particular, the input to the first layer, instead of being the set of encoder features \(\mathbf{F}\), becomes \(\tau_1\left( \mathbf{F} \right)\); similarly, subsequent layers perform a similar operation on feature extracted from the second branch, such that:
\begin{equation}
\bm{\gamma}_i = g_i \left( \bm{\gamma}_{i-1} \right) \oplus \tau_i \left( \bm{\phi}_i \right) .
\end{equation}
The reason for integrating features \(\bm{\Phi}\) from the second branch into \( \bm{\Gamma} \) is twofold: first, \(\bm{\Phi}\) features internally encode global information from \(\bm{\Theta}\), which ought to be taken into account to compensate for the removal of the temporal dimension; second, since the third branch focuses on spatial relations, it makes sense to provide it with the largest resolution available at that stage.

Finally, output features from each branch are processed by a late fusion layer to produce the final saliency map \(\mathbf{S}\) as:
\begin{equation}
 \mathbf{S}= o\left( \left[ \sigma_n\left( \bm{\theta}_N , \bm{\phi}_N , \bm{\gamma}_N \right] \right) \right),
\end{equation}
with $o$ being the transformation applied by the fusion layer and \(\left[ \cdot \right]\) denoting the concatenation operator.

\subsection{Training Objective}
Given an input video sequence \(\textbf{V} \in \mathbb{R}^{T \times H \times W \times 3}\) and the ground-truth saliency map \( \textbf{G} \in \mathbb{R}^{H \times W}\) for the last frame of the clip, the objective of the model is to estimate a saliency map \(\textbf{S} \in \mathbb{R}^{H \times W}\) for the corresponding frame. Our training objective $\mathcal{L}$ is inspired from~\cite{TMFI} and defined as follows:

\begin{equation}
\mathcal{L}(\textbf{S}, \textbf{G}) = \mathcal{L}_\text{KL}( \textbf{S}, \textbf{G}) + \mathcal{L}_\text{CC}(\textbf{S}, \textbf{G}) 
\label{eq:m}
\end{equation} 

The \(\mathcal{L}_\text{KL}\) loss term treats the predicted and ground-truth saliency maps as two probability distribution, and estimates their distance by means of the Kullback-Leibler divergence:
\begin{align}
\mathcal{L}_\text{KL}(\textbf{S}, \textbf{G}) = \sum\limits_{x} \textbf{G}(x) \log \frac{\textbf{G}(x)}{\textbf{S}(x)}
\end{align}
where \(x\) scans pixel locations.

The \(\mathcal{L}_\text{CC}\) loss term computes the correlation coefficient between the saliency maps, considering them as random variables:

\begin{align}
\mathcal{L}_\text{CC}(\textbf{S},\textbf{G}) = - \frac{\text{cov}(\textbf{S},\textbf{G})}{\rho(\textbf{S}) \rho(\textbf{G})}
\end{align} 

where \(\text{cov}(\textbf{S},\textbf{G})\) is the covariance of \(\textbf{S}\) and \(\textbf{G}\) and \(\rho(\cdot)\) is the standard deviation operator.

\section{Experiments}
In this section, we present our comprehensive experiments designed to showcase the superiority of our proposed model, SalFoM. We specifically describe three benchmark datasets for Video Scene Parsing (VSP) in Section~\ref{dataset}. Following that, we elaborate on the experimental setup, analyze the results, and conduct an ablation study in Sections~\ref{setup}, \ref{Results}, and \ref{ablation}, respectively.

\subsection{Datasets}\label{dataset}

We extensively evaluate our model's performance on three commonly used benchmark datasets: DHF1K~\cite{DHF1K}, UCF-Sports~\cite{UCF}, and Hollywood-2~\cite{Hollywood}. DHF1K, the largest eye-tracking dataset for dynamic fixation prediction, consists of 1,000 annotated videos divided into train (600), validation (100), and test (300) sets. Ground truths for the test set are not released, so quantitative evaluations are provided by the dataset's curators.
The Hollywood-2 dataset~\cite{Hollywood} contains 1,707 video clips extracted from 69 Hollywood movies, grouped by 12 action categories. 
Unlike the DHF1K dataset, this dataset employs a task-driven viewing approach for video annotation. The annotations were gathered from three different perspectives: context recognition by four observers, free viewing by three observers, and action recognition by twelve observers. The UCF-Sports dataset~\cite{UCF}, which is derived from the UCF sports action dataset, consists of 150 videos spanning nine sports classes. It is annotated using the same task-driven viewing methodology as the Hollywood-2 dataset. Following the approach in \cite{STSANet}, we utilized 103 videos for training and 47 videos for testing.

\subsection{Experimental Setup}\label{setup}

When training on DHF1K, we initialize the encoder of our network from the pretrained weights from UMT-L/16 on the Kinetics-400 dataset~\cite{k400}; when training on Hollywood-2 and UCF-sports, instead, we initialize the encoder using weights obtained after training on DHF1K. In both cases, encoder parameters are fine-tuned at training time. We train our model using a batch size of 1, and employing the Adam optimizer~\cite{kingma2014adam} for gradient descent, with an initial learning rate of $10^{-5}$.

At each training iteration, the network processes 16 consecutive video frames with a spatial resolution of 224$\times$224, and predicts the saliency map for the last frame of the video sequence. We implement early stopping based on the performance on the target dataset's validation set.

At inference time, we generate saliency maps for all video frames using a sliding window approach, as utilized by~\cite{TASED-Net}. To ensure sufficient temporal context for the initial frames of a clip, we reverse the order of the frames. To assess our network's performance, we used three location-based metrics, Shuffled AUC (S-AUC), AUC-Judd (AUC-J), and Normalized Scanpath Saliency (NSS), as well as two distribution-based metrics, Linear Correlation Coefficient (CC) and Similarity Metric (SIM)~\cite{bylinskii2018different}.

\begin{table*}[t]
\centering
\caption{Quantitative comparison of different models on DHF1K dataset. The top score in each metric is in bold.}
\begin{tabular}{lcccccr}
\toprule
\textbf{Models} & \multicolumn{5}{c}{\textbf{DHF1K}} \\
\midrule
& AUC-J & SIM & S-AUC & CC & NSS  \\
\midrule
SalEMA & 0.890 & \textbf{0.466} & 0.667 & 0.449 & 2.574 \\
STRA-Net & 0.895 & 0.355 & 0.663 & 0.458 & 2.558 \\
TASED-Net & 0.895 & 0.361 & 0.712 & 0.470 & 2.667 \\
SalSAC & 0.896 & 0.357 & 0.697 & 0.479 & 2.673 \\
UNISAL & 0.901 & 0.390 & 0.691 & 0.490 & 2.776 \\
ViNet & 0.908 & 0.381 & 0.729 & 0.511 & 2.872 \\
HD2S & 0.908 & 0.406 & 0.700 & 0.503 & 2.812 \\
VSFT & 0.910 & 0.410 & 0.720 & 0.518 & 2.977 \\
TSFP-Net & 0.911 & 0.392 & 0.723 & 0.516 & 2.966 \\
STSANet & 0.912 & 0.382 & 0.722 & 0.528 & 3.010\\
THTD-Net & 0.915 & 0.406 & 0.729 & 0.547 & 3.138 \\
TMFI-Net & 0.915 & 0.406 & 0.730 & 0.546 & 3.146\\
\midrule
\textbf{SalFoM (Ours)} &  \textbf{0.922} & 0.420  & \textbf{0.735} &  \textbf{0.569} & \textbf{3.353}  \\
\bottomrule
\end{tabular}
\label{tab:results}
\end{table*}

\begin{figure*}[t]
    \centering
    \includegraphics[width=0.94\textwidth]{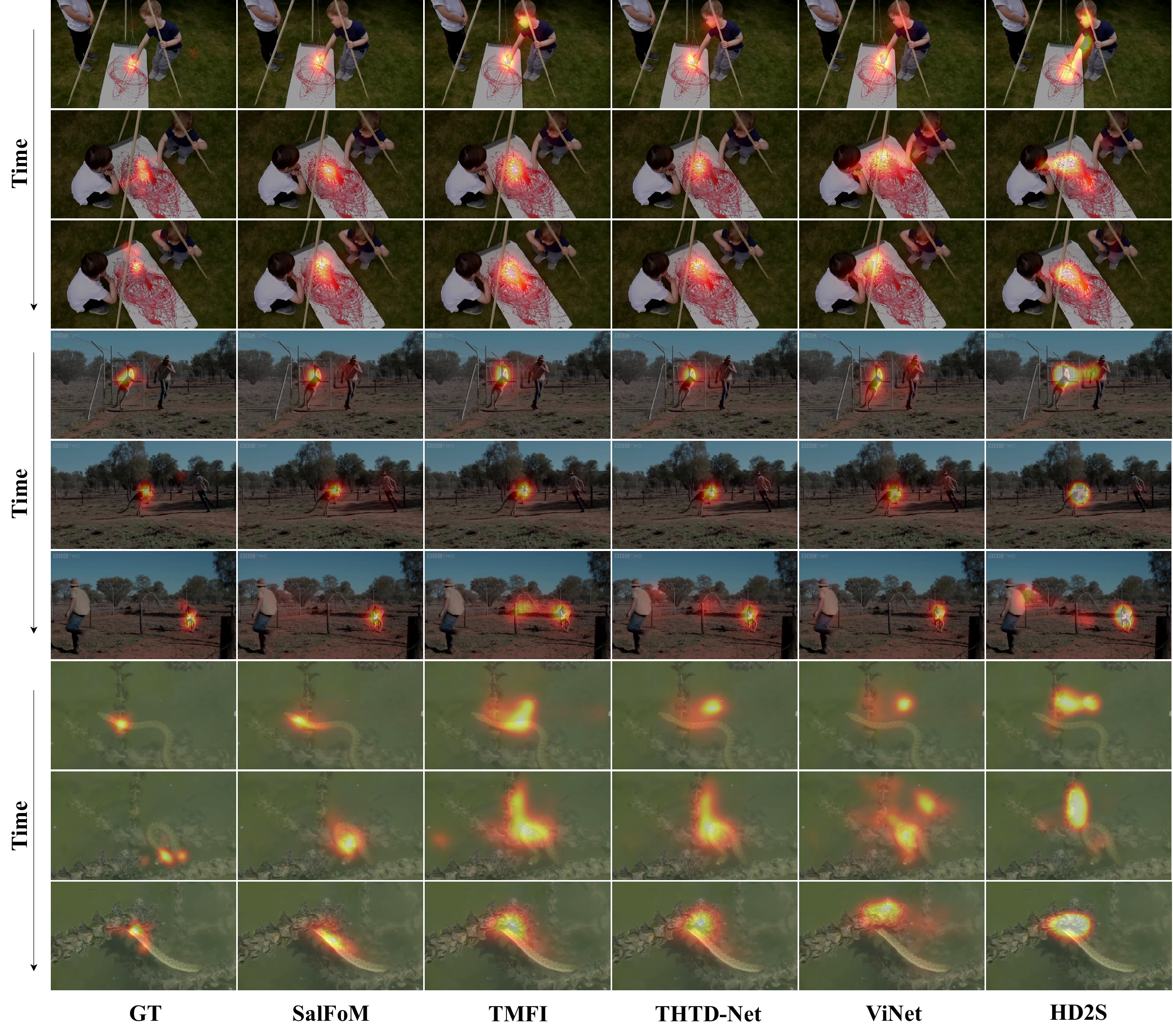}
    \caption{Qualitative comparison of the performance of different VSP models on DHF1K.}
    \label{Temporal_maps}
\end{figure*}

\begin{figure*}[t]
    \centering
    \includegraphics[width=1\textwidth]{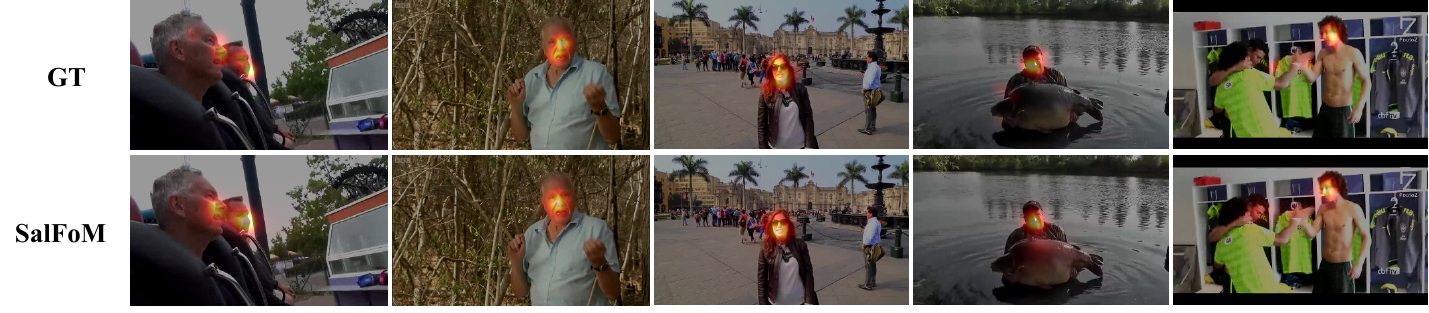}
    \caption{Qualitative evaluation of the performance of SalFoM for predicting saliency of faces against ground truths, on DHF1K.}
    \label{face}
\end{figure*}

\begin{table*}[t]
\centering
\caption{Quantitative comparison of different models on Hollywood-2 and UCF-Sports datasets. The top score in each metric is in bold.}
\begin{tabular}{lcccccccc}
\toprule
\textbf{Models} & \multicolumn{4}{c}{\textbf{Hollywood-2}} & \multicolumn{4}{c}{\textbf{UCF-Sports}} \\
\cmidrule(r){1-1} \cmidrule(lr){2-5} \cmidrule(lr){6-9}
& AUC-J & SIM & CC & NSS & AUC-J & SIM & CC & NSS \\
\cmidrule(r){1-1} \cmidrule(lr){2-5} \cmidrule(lr){6-9}
SalEMA & 0.919 & 0.487 & 0.613 & 3.186 & 0.906 & 0.431 & 0.544 & 2.638 \\
STRA-Net  & 0.923 & 0.536 & 0.662 & 3.478 & 0.910 & 0.479 & 0.593 & 3.018 \\
TASED-Net  & 0.918 & 0.507 & 0.646 & 3.302 & 0.899 & 0.469 & 0.582 & 2.920 \\
SalSAC  & 0.931 & 0.529 & 0.670 & 3.356 & 0.926 & 0.534 & 0.671 & 3.523 \\
UNISAL  & 0.934 & 0.542 & 0.673 & 3.901 & 0.918 & 0.523 & 0.644 & 3.381 \\
ViNet  & 0.930 & 0.550 & 0.693 & 3.73 & 0.924 & 0.522 & 0.673 & 3.62\\
HD2S  & 0.936 & 0.551 & 0.670 & 3.352 & 0.904 & 0.507 & 0.604 & 3.114 \\
VSFT & 0.936 & 0.577 & 0.703 & 3.916 & - & - & - & -\\
TSFP-Net & 0.936 & 0.571 & 0.711 & 3.910 & 0.923 & 0.561 & 0.685 & 3.698 \\
STSANet & 0.938 & 0.579 & 0.721 & 3.927 & \textbf{0.936} & 0.560 & 0.705 & \textbf{3.908} \\
THTD-Net & 0.939 & 0.585 & 0.726 & 3.965 & 0.933 & \textbf{0.565} & \textbf{0.711} & 3.840 \\
TMFI-Net & \textbf{0.940} & \textbf{0.607} & \textbf{0.739} & \textbf{4.095} & \textbf{0.936} & \textbf{0.565} & 0.707 & 3.863\\
\cmidrule(r){1-1} \cmidrule(lr){2-5} \cmidrule(lr){6-9}
\textbf{SalFoM (Ours)}&0.935&0.583&0.709&3.902&0.928&0.516&0.631&3.543\\
\bottomrule
\end{tabular}
\label{tab:H_U_results}
\end{table*}

\begin{table}
\centering 
\caption{Ablation study: assessing the impact of SalFoM model components on validation set of DHF1K.}
\begin{tabular}{lcccc}
\toprule
\textbf{Model} & \textbf{CC} & \textbf{NSS} & \textbf{SIM} & \textbf{AUC-J}\\
\midrule
Encoder: VidSwin-S  & 0.512 & 2.917 & 0.379 & 0.916 \\
Encoder: UMT-L (8 Frames) & 0.552 & 3.169 & 0.418 & 0.924 \\ 
Encoder: UMT-B (8 Frames) & 0.527 & 2.935 & 0.400 & 0.915 \\ 
\midrule
Decoder: single TCFE branch & 0.560 & 3.245 & 0.433 & 0.926 \\ 
Decoder: single DFD branch & 0.564 & 3.288 & 0.425 & 0.926 \\ 
Decoder: single SFD branch & 0.560 & 3.220 & 0.408 & 0.926 \\ 
\midrule
Decoder: SFD + DFD branches & 0.564 & 3.291 & 0.429 & 0.927 \\ 
Decoder: SFD + TCFE branches & 0.563 & 3.240 & 0.415 & 0.918 \\ 
Decoder: DFD + TCFE branches & 0.564 & 3.294 & 0.433 & 0.927 \\ 
\midrule
\textbf{SalFoM (ours)} & \textbf{0.565} & \textbf{3.299} & \textbf{0.436} & \textbf{0.928} \\
\bottomrule
\end{tabular}
\renewcommand\thetable{2}
\label{tab:ablation}
\end{table}

\begin{figure*}[h!]
    \centering
    \includegraphics[width=0.8\textwidth]{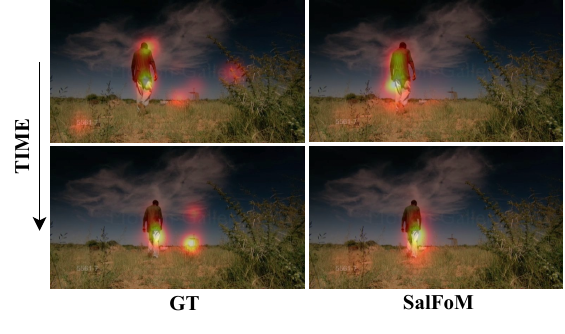}
    \caption{A sample of failure case on UCF-Sports dataset, due to the task-driven annotation methodology.}
    \label{UCF}
\end{figure*}

\subsection{Result Analysis}\label{Results}

We quantitatively compare the performance of our model with state-of-the-art VSP models that have demonstrated the best results on the DHF1K benchmark\footnote{https://mmcheng.net/videosal}: SalEMA~\cite{linardos2019simple}, STRA-Net~\cite{lai2019video}, TASED-Net~\cite{TASED-Net}, SalSAC~\cite{salsac}, UNISAL~\cite{UNISAL}, ViNet~\cite{ViNet}, HD2S~\cite{HD2S}, VSFT~\cite{ViNet}, TSFP-Net~\cite{TSFP-Net}, STSANet~\cite{STSANet}, TMFI-Net~\cite{TMFI}, and THTD-Net~\cite{THTD}. The results presented in Table~\ref{tab:results} demonstrate that on the DHF1K dataset, SalFoM outperforms other state-of-the-art models across almost all evaluation metrics (the only exception being SalEMA on SIM, although it performs significantly worse on the other metrics), notably surpassing the top-ranked TMFI-Net model. Moreover, qualitative examples reported in Figure~\ref{Temporal_maps} confirm that the proposed approach, when compared with state-of-the-art VSP models, aligns better with ground-truth annotations, focusing on single points of interest, whereas other methods distribute their attention towards less salient portions of the video. Finally, as proposed by Bylinskii et al.~\cite{bylinskii2016should}, evaluating the capability and performance of saliency prediction models in identifying high-level concepts, such as face recognition, in visual scenes can serve as an indicator of their effectiveness. We qualitatively perform such analysis and report some samples in Figure~\ref{face}, again showing the alignment between our model's predictions and the ground truth.

Results on Hollywood-2 and UCF-Sports are presented in Table~\ref{tab:H_U_results}. We can see that our model achieves comparable results to state-of-the-art methods on the Hollywood-2 dataset, while its performance drops in certain metrics when assessed on the UCF-Sports dataset. This pattern of performance drop is not unique to our model but is also observed in other leading methods, and its causes can be led to certain characteristics of those datasets~\cite{HD2S}, including task-driven observations and center bias.  As illustrated in Figure~\ref{UCF}, our model, SalFoM, effectively detects the human subject, representing the most salient region, yet it does not precisely match the ground truth. This discrepancy arises because the annotations accompanying the Hollywood-2 and UCF-Sports datasets primarily emphasize actions rather than salient regions. Additionally, the lower performance on these datasets can be attributed to the scarcity of diverse spatio-temporal data, particularly notable in Hollywood-2, which comprises numerous short videos.
An additional challenge arises from the relatively small size of the UCF-Sports dataset when compared to DHF1K and Hollywood-2. In our case, the limited size of the dataset contributes to overfitting of the model and negatively impacts its ability to generalize. Additional model evaluations on DHF1K are reported in the supplementary materials.

\subsection{Ablation Study}\label{ablation}

In this section, we assess the impact of various components within our model by designing and evaluating different model variants on DHF1K, utilizing the validation set as the test set. The results are reported in Table~\ref{tab:ablation}. We first explore the influence of the encoder employed, by replacing UMT-L/16 with a Video Swin Transformer, similarly to TMFI-Net, and with variants of UMT that process 8 frames instead of 17. We accordingly configure our model's decoder parameters to align with the different temporal sizes of the encoded features. Results show that non-VFM-based encoders fail to yield satisfactory features, and that reducing the input frames of a VFM degrades performance.

After showing the superiority of the UMT-based feature encoder, we carry out additional experiments to demonstrate the importance of the proposed decoder strategy. In particular, we evaluate the performance of variants of our decoder which use either a single or a combination of two (out of three) branches. While all configurations are able to achieve satisfactory results, the full combination of all three decoding branches yields the best values for the employed metrics.

\section{Conclusion}

In this work, we present SalFoM, a video saliency prediction model that incorporates UMT, an innovative video foundation model, into our network architecture. This integration enables the model to effectively capture both temporal (scene dynamics) and spatial (object-related) information. For the decoder component of SalFoM, we introduce a three-branch structure that includes a locality-aware spatio-temporal transformer branch, a branch based on 3D convolutional layers, and another based on 2D convolutional layers. This configuration is designed to merge local and global spatio-temporal signals from diverse perspectives to generate the final saliency map. Our experiments demonstrate that SalFoM outperforms existing VSP models on several standard evaluation metrics.

\bibliographystyle{splncs04}
\bibliography{main}
\end{document}